%% file: main.tex
\documentclass{article}

    \usepackage[final,nonatbib]{ml4mols_2020}

\usepackage[numbers]{natbib}

\usepackage[utf8]{inputenc} 
\usepackage[T1]{fontenc}    
\usepackage{hyperref}       
\usepackage{url}            
\usepackage{booktabs}       
\usepackage{amsfonts}       
\usepackage{nicefrac}       
\usepackage{microtype}      
\usepackage{epsfig}
\usepackage{amsmath,longtable,fancyhdr,booktabs,multirow,graphicx,subfigure,float}
\usepackage{amssymb}
\usepackage{amsmath}
\usepackage{amsthm}
\usepackage{amssymb}
\usepackage{bbm}
\usepackage{mathtools}
\usepackage{mathrsfs}
\usepackage{dsfont}
\mathtoolsset{showonlyrefs}
\usepackage{caption}
\usepackage{algorithm}
\usepackage{algorithmic}
\usepackage{enumitem}
\usepackage{hyperref}
\usepackage{times}
\usepackage{xcolor}
\usepackage{hyperref}
\usepackage{tikz}
\usepackage{physics}
\usepackage{mathdots}
\usepackage{yhmath}
\usepackage{cancel}
\usepackage{color}
\usepackage{siunitx}
\usepackage{array}
\usepackage{multirow}
\usepackage{gensymb}
\usepackage{wrapfig}
\usepackage{lipsum}
\usepackage{makecell}
\usepackage{nccmath}
\usepackage[export]{adjustbox}
\usepackage{tabularx}
\usepackage{bigstrut, multirow, makecell, diagbox}
\usetikzlibrary{fadings}
\usetikzlibrary{patterns}
\usetikzlibrary{shadows.blur}
\usetikzlibrary{shapes}

\usepackage{appendix}

\usepackage{theoremref}

\newcommand{\be}{\begin{equation}}
    \newcommand{\ee}{\end{equation}}

\makeatletter
\DeclareRobustCommand\bfseriesitshape{%
  \not@math@alphabet\itshapebfseries\relax
  \fontseries\bfdefault
  \fontshape\itdefault
  \selectfont
}
\makeatother
\DeclareTextFontCommand{\textbfit}{\bfseriesitshape}

\newcommand{\xx}{\mathbf{x}}

\newcommand{\MLP}{\text{MLP}}

\newcommand{\bfA}{\mathbf{A}}

\newcommand{\Eb}{{\mathbb{E}}}
\newcommand{\En}{{\mathcal{E}}}

\def\MODEL{ASTRAGEM}

\newcommand{\isep}{\mathrel{{.}\,{.}}\nobreak}

\usepackage{xcolor}
\definecolor{dark-red}{rgb}{0.4,0.15,0.15}
\definecolor{dark-blue}{rgb}{0,0,0.7}
\hypersetup{
    colorlinks, linkcolor={dark-blue},
    citecolor={dark-blue}, urlcolor={dark-blue}
}

\title{Adversarial Stein Training for Graph Energy Models}

\author{%
  Shiv Shankar\\
  University of Massachussets\\
  \texttt{sshankar@cs.umass.edu} \\
}

\begin{document}

\maketitle
\begin{abstract}
Learning distributions over graph-structured data is a challenging task with many applications in  biology and chemistry. In this work we use an energy-based model (EBM) based on multi-channel graph neural networks (GNN) to learn permutation invariant unnormalized density functions on graphs. Unlike standard EBM training methods our approach is to learn the model via minimizing adversarial stein discrepancy. Samples from the model can be obtained via Langevin dynamics based MCMC. We find that this approach achieves competitive results on graph generation compared to benchmark models.

\end{abstract}

\section{Introduction}

Modeling and generating molecular structures is of great value in applications such as drug discovery \cite{bonetta2019}  and immunology \cite{Crossman2020}. Graphs provide a useful mathematical abstraction to express and analyse molecular structures. They have also been frequently used to captures relational structure in language processing\citep{hamaguchi2017knowledge}, systems \citep{batagelj2003m} and bio-chemistry\cite{bonetta2019} . Generative models for graphs also have applications in network analysis \cite{albert2002statistical}, chemical analysis \cite{PCSJohn2019} and other fields. 

Recently there have been a number of works on learning generative models of graphs from data. One framework for such models is latent variable based autoencoders. This framework includes models such as GraphVAE~\citep{simonovsky2018graphvae} and junction tree variational autoencoders~\citep{jin2018junction}. These models typically use a graph neural networks (GNN) ~\citep{gori2005new,scarselli2008graph} to encode the graph into a latent space, and generate samples by decoding latent variables sampled from a prior distribution. Another paradigm for graph generation is grow the graph one node (or one subgraph) at a time.  ~\citet{li2018learning,you2018graph,liao2019lanczosnet} take such an autoregressive approach, where graphs are generated sequentially

\label{sec:intro}

Our work focuses on using Energy based models (EBM) to model permutation invariant distributions on graphs. Unlike the earlier mentioned auto-encoding and autoregressive approaches, energy-based model provides an extremely flexible way to model densities. This allows one to naturally enforce desirable inductive biases and incorporate domain knowledge into the models. For example one can write models which by construction are permutation invariant by choosing appropriate energy function. EBMs and similar global unnormalized models have long been used for schema-modeling \cite{deshpande2007probabilistic,godbole2004discriminative} and structure prediction  \cite{andor2016globally}. They have also shown promise in applications like speech processing~\cite{wang2018learning}~\nocite{torkamani2020differential} and protein-folding~\citep{du2020energy,ingraham2018learning}.


In this work we utilize GNNs to construct a permutation invariant EBM for distribution over graphs. More specifically we used the learnable multi-channel approach used in \cite{niu2020permutation} to design our energy function. Then we rely on using modification of Stein's Identity \cite{stein1972bound} to adversarially train EBMs without the need to sample which allows scaling to larger datasets.

\section{Preliminaries}

A graph $G$ is a tuple of finite set of nodes $V_G$ and a finite set of edges $E_G$ which connect the nodes. Each edge $e$ is identified by a node pair $(u,v)$. A graph $G$ is said to be undirected if its edges are be undirected i.e. if $(u,v) \in E_G$ then its implied that $(v,u) \in E_G$. Any unweighted graph can be represented by a symmetric matrix $\bfA \in \{0,1\}^{|V_G|\cross|V_G|}$ called its adjacency matrix.  We shall use the adjacency matrix representation for our graphs. In general the adjacency matrix representation $\bfA$ of a graph depends on the ordering of nodes. However in this paper we will be dealing only with permutation invariant functions. A distribution of graphs is then equivalent to a distribution of adjacency matrices $p(\bfA)$.

\subsection{Energy Based Models}

Energy-based models (EBMs) assign to each point $\xx$ in the input space $\Omega$ an unnormalized log probability $\En_\theta(\xx)$.  EBMs get their name from the so-called energy function, which is simply $\En$. This function fully specifies the distribution over the data as :
$$p_\theta(x) = \exp(\En_\theta(\xx))/Z(\theta)$$

Here, $\theta$ represents the model parameters, and $Z(\theta) = \sum_{x \in \Omega} exp(\En_\theta(x))$ is the normalization constant. Because $Z(\theta)$ is typically intractable, most EBMs cannot be trained by maximum likelihood and instead must rely on alternatives relying on MCMC sampling; which is unscalable for large datasets. There are also sampling-free methods to train non-normalized models. We refer to Appendix \ref{apx:relword_ebm} for discussion of such methods.

\subsection{Stein Discrepancy}

Stein Discrepancy~\citep{gorham2015measuring} between two distributions $p$ and $q$ is given by:

\begin{align*}
S(q,p) = \max_{f \in \mathcal{F}} \Eb_{q(z)}[ (\partial_z \log p(z))^Tf(z) + Tr(\partial_z f(z))]
\end{align*}
where $\mathcal{F}$ is the family of function to over which the maximization is done. This was proposed as an extension Stein's Identity \citep{stein1972bound} and it can be shown that if $\mathcal{F}$ is a sufficiently rich class of functions, then the above expression is indeed a divergence ~\citep{gorham2015measuring}. Note that the terms inside the expectation can be computed even for an unnormalized distribution $p$ since $\partial_z \log p(z)$  does not depend on the normalization constant of $p$. 



\section{Adversarial Stein Training for Graph Energy Models}
In this section  we present our approach for modeling graph distributions dubbed \MODEL \space ( \textbfit{A}dversarial \textbfit{S}tein \textbfit{Tra}ining for \textbfit{G}raph \textbfit{E}nergy \textbfit{M}odels ). Our approach consists of three components. First using an Edgewise Dense Prediction Graph Neural Network (EDPGNN) \citep{niu2020permutation} architecture we write a powerful permutation invariant energy function. Second we use a novel approach (labeled Adversarial Stein Training) to learn such energy models without requiring computationally burdensome sampling during training.

We utilize the idea of estimating the Stein Discrepancy \cite{gorham2015measuring} via optimization over parametric function spaces. We restrict $\mathcal{F}$ to be the space of functions computable by a specific neural network architecture $C$ which is parameterized by $\psi$.
The discrepancy measured by a specific instance of the network $C_\psi$

\begin{align*}
S(q,p, \psi) = \Eb_{q(z)}[ (\partial_z \log p(z))^TC_\psi(z) + Tr(\partial_z C_\psi(z))]
\end{align*}

A quantitative value of the Stein discrepancy $\hat{S}$ can then be obtained via maximizing $S(q,p, \psi)$. From the perspective of implementation we perform this optimization via gradient descent
\begin{align*}
\hat{S}(q,p) = \max_\psi \Eb_{q(z)}[ (\partial_z \log p(z))^TC_\psi(z) + Tr(\partial_z C_\psi(z))]
\end{align*}

Note that computing the above discrepancy only requires samples from $q$. Furthermore computing the $\partial_z \log p(z)$ for an EBM is simple as:
$$ 
\partial_z \log p(z) = \partial_z \log \exp(\En_\theta(z))/Z(\theta) = \partial_z \En_\theta(z)
$$
is independent of $Z$ and directly computable via the energy function $\En_\theta$. 
Once the discrepancy is estimated we can minimize the same over the parameters of our energy model $\theta$. The entire procedure is analogous to training a GAN \citep{arjovsky2017wasserstein}, hence the name adversarial training. The discrepancy estimator $C_\psi$ can be thought of as an adversary or critic, while the generator is the energy model $\En_\theta$. As such we will sometimes refer to the discrepancy estimator as a "critic". For our current purpose of learning an energy model over graphs, we set $q$ to be the empirical distribution of graphs from the data. The model probability $p$ is chosen to be an energy model given by an EDPGNN (parameterized via $\theta$); which makes our distribution inherently permutation invariant. Adversarially learnt models often produce realistic but non-diverse samples~\citep{arora2018do}, however incorporating domain knowledge into the adversary can help alleviate such issues~\citep{xu2018dp}. Our approach allows us to bring domain insights into both the generator $\En_\theta$ and the critic  $C_\psi$


\subsection{Noise Smoothening and Critic Regularization}
Since the empirical distribution $q$ over graphs is discrete while the $p_\theta$ is continuous the ideal critic can easily distinguish the two leading to large gradients and unstable training. In our experiments clipping or norm-regularizing the critic parameters $\psi$ were not sufficient to alleviate the problem. As such to stabilize the training we employed multiple tricks. First in each minibatch we add Gaussian noise $ \mathcal{N}(0,\sigma^2) $ to the adjacency matrices of the graphs and symmetrize them \footnote{\textbf{SYM} refers to symmetrization operation} before using the matrices as inputs. Secondly instead of using a single noise variance we used multiple values of $\sigma^2$ at the same time during training to get different corruptions of the same graph. Thirdly instead of a single critic we use a series of noise conditioned critics $C^i_{\psi,\sigma_i}$ which are regularized by parameter sharing.
Finally we add extra regularization to the critics by trying to reduce their power against a Gaussian kernel-density estimate $p^{KDE}_{h}(\xx)$ of the distribution. More details of the exact procedure are available in the Appendix \ref{apx:reg}.

The overall training procedure is presented in Algorithm \ref{alg:nsd_train}

\begin{algorithm}
\caption{Adversarial Stein Training for Graph Energy Model}\label{alg:nsd_train}
\begin{algorithmic}[h]
\REQUIRE Critic architecture $C_{\psi}$, EBM architecture $\En_\theta$, data $ D = \{\bfA_j\}_{j=1}^n$, noises $\{\sigma_i\}_{i=1}^k$\\
\REQUIRE Hyperparameters: Regularization $\lambda_K,\lambda_{L_2}$, Critic Update Steps $\text{C}_\text{iter}$, Total Iterations $T$

\FOR{$T$ iterations}
   \STATE Sample $\bfA^G$ from $D$
   \STATE $q^G_i = \textbf{SYM}( \mathcal{N}(\bfA^G, \sigma_i^2)) \space \forall i \in [1 \isep k] $
   
   \FOR{$\text{C}_\text{iter}$ iterations }
   \STATE Sample $\bfA^C$ from $D$
   \STATE $q^{Cr}_i = \textbf{SYM}( \mathcal{N}(\bfA^C, \sigma_i^2)) \space \forall i \in [1 \isep k] $

   \STATE Update $\psi$ with $\nabla_\psi \sum_i  \left( S(q^{Cr}_i, \En_\theta,C_{\psi,\sigma_i}) - \lambda_K
    |S(q^{Cr}_i, p^{KDE}_{\sigma_i},C_{\psi,\sigma_i})| \right)  -  \lambda_{L_2}|\psi|^2
   $
   \ENDFOR
   
   \STATE Update $\theta$ with $-\nabla_\theta \sum_i \left( S(q^G_i, \En_\theta,C_{\psi,\sigma_i}) \right)$

\ENDFOR
\STATE Return resulting model $\En_\theta$
\end{algorithmic}
\end{algorithm}

\subsubsection{Graph Sampling} 
Once we have a trained energy model, we use the following procedure to sample new graphs from the distribution. We first obtain the number of nodes $N$ in the graph. For this the approach of \citet{ziegler2019latent, niu2020permutation} is taken which samples from the empirical multinomial distribution of node sizes in the training data. Once $N$ is fixed, we can sample matrix $\bfA \in \mathbb{R}^{N\times N}$ via Langevin dynamics on the energy function $\En_\theta$. 

\section{Experiments}

We present empirical findings regarding our training method here. The results show this method can be used to learn good-quality generative models.

\paragraph{Community Graph Learning}
We train EBMs on the two community graph datasets used commonly for graph learning \citet{you2018graphrnn} :
\begin{itemize}
    \item Ego-small: 200 graphs with $4 \leq |V| \leq 18$, obtained from the Citeseer network.
    \item Community-small: 100 two-community graphs with $12 \leq |V| \leq 20 $ that were generated procedurally, 
\end{itemize}
We compare our approach against other recent generative models like GraphRNN \citep{you2018graph}, GraphVAE \citep{simonovsky2018graphvae} and DeepGMG \citep{li2018learning}.
The evaluation method followed is the same as the one prescribed in \citet{you2018graph} which computes MMD scores between generated samples and the test set for 3 graph statistics: degree, orbit,and cluster. Our results are summarized in 
Table \ref{tab:MMD} where we can see our method outperforms other methods on most metrics. 

\begin{table*}[t]
\centering
\begin{adjustbox}{max width=0.85\textwidth}
\begin{tabular}{@{}lllllllll@{}}
\toprule
\multirow{2}{*}{Model} & \multicolumn{4}{c}{Community-small}                               & \multicolumn{4}{c}{Ego-small}                                      \\ \cmidrule(lr){2-9}
                       & Deg.           & Clus.          & Orbit          & Avg.           & Deg.           & Clus.          & Orbit          &  Avg.                      \\ \midrule
GraphVAE               & 0.350          & 0.980          & 0.540          & 0.623          & 0.130          & 0.170          & 0.050          & 0.117                           \\
DeepGMG                & 0.220          & 0.950          & 0.400          & 0.523          & \textbf{0.040}          & 0.100          & 0.020          & 0.053                           \\
GraphRNN               & 0.080          & \textbf{0.120} & 0.040          & 0.080          & 0.090          & 0.220          & 0.003          & 0.104                           \\
\MODEL             & \textbf{0.050} & 0.124          & \textbf{0.026} & \textbf{0.067} & 0.045          & \textbf{0.090} & \textbf{0.006}          & \textbf{0.047}                  \\ 
\bottomrule
\end{tabular}
\end{adjustbox}

\caption{MMD results of various graph generative models}
\label{tab:MMD}
\end{table*}

\paragraph{Molecule Generation}
The goal of molecule generation is to learn a distribution of valid molecules from a database of molecular structures.
A molecule can be represented as a graph with the atoms becoming nodes, and the atomic bonds becoming edges.  All the nodes and edges have associated categorical information about the type and nature of the atoms and bonds respectively. We test the performance of \MODEL \space on the Zinc250k dataset \citep{irwin2012zinc}. 

\begin{figure*}[ht]
    \centering
	\subfigure[Training data]{
        \includegraphics[width=.45\linewidth, trim={0 0 0 1cm},clip]{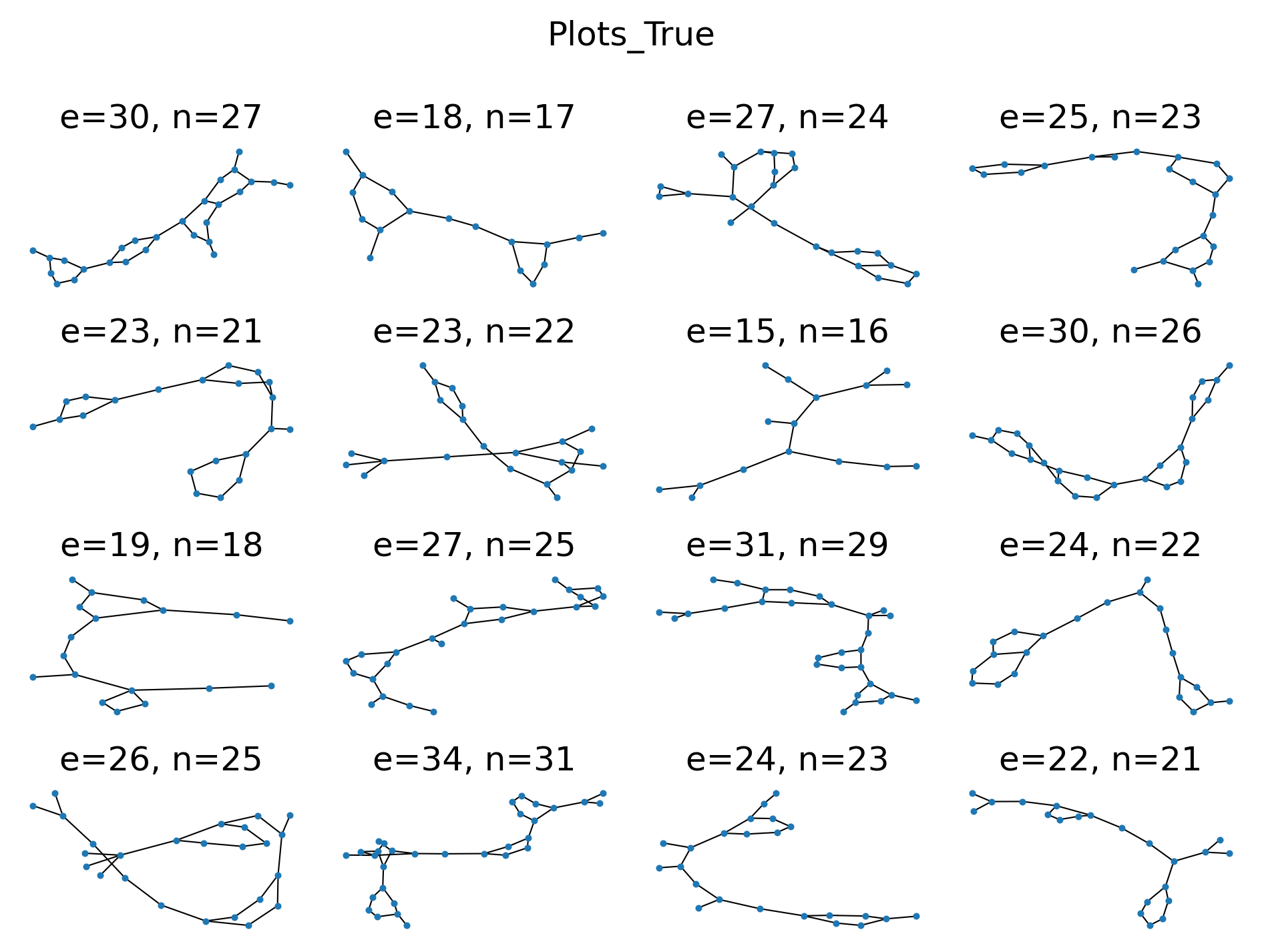}
    } \quad
    \subfigure[\MODEL \space samples]{
        \includegraphics[width=.45\linewidth, trim={0 0 0 1cm},clip]{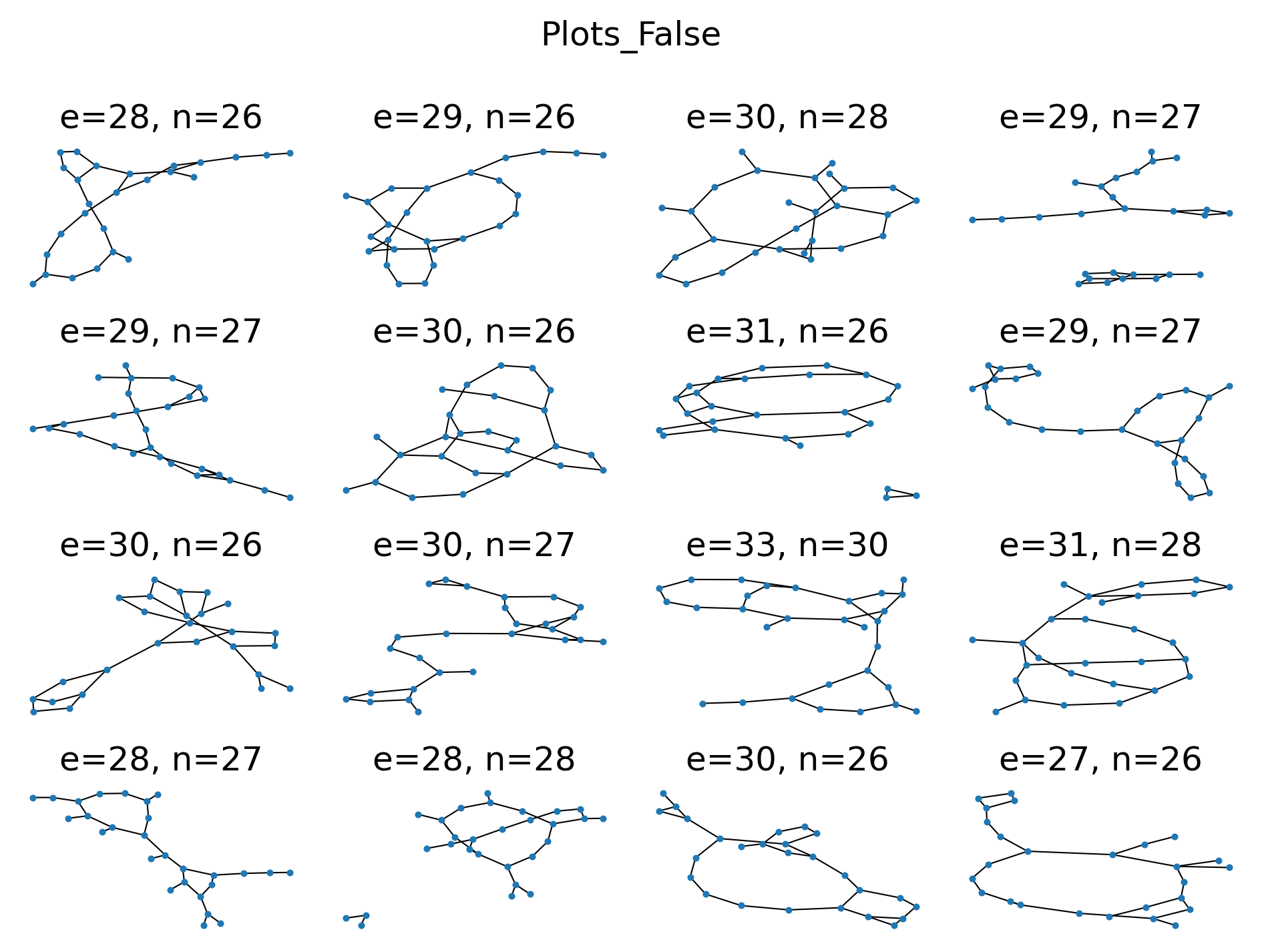}
    }
\caption{Samples from the training data and \MODEL \space on the Zinc250k dataset \citep{irwin2012zinc}}
\label{fig:ZincSamples}
\end{figure*}

Our current approach only learns the graph structure and not the node and edge labels which makes comparison with existing works on the dataset difficult. As such we refrain from making any comparisons and instead present some samples as generated by our model in Figure \ref{fig:ZincSamples}.  

\section{Conclusion}

This work explored an adversarial training approach based on Stein Divergence to learn an EBM for model graph distributions. The results show our method has potential to be a useful generative models for molecular graphs. Our method can be considered a version of adversarial inference, which opens up possibilities of bringing other advances in the field \cite{dumoulin2016adversarially,shankar2020bosonic} into learning generative models from graphs. 
Furthermore, while these preliminary results seem promising, more research is needed to successfully use our model for generation of full molecular graphs i.e all atom and bond features. 


\bibliography{main}
\bibliographystyle{abbrvnat}

\clearpage
\input{appendix1.tex}

\end{document}

%% file: appendix1.tex
\appendix

\onecolumn

\section{Noise Smoothening and Regularization}
\label{apx:reg}
Since the optimal critic $C$ for a pair of distributions $p,q$ need not be optimal for another pair, instead of a single critic using multiple critics was far more effective. We use a series of critics $C^i_{\psi}$ that are conditioned on their corresponding noise levels. These are regularized by sharing parameters across the critics. For this purpose we use the following structure for the layers in the critic. Each MLP layer $f_j$ in the noise conditional critic $C^i_{\psi}$ is given as
\begin{displaymath}
 	f_{j,i}(\bfA) = \operatorname{Activation}((\mathbf{W}_j\bfA + \mathbf{b}_j)\alpha_i + \beta_i)
\end{displaymath}
where $\alpha_i, \beta_i$ are individual parameters for each noise level $\sigma_i$ and $\mathbf{W}_j,\mathbf{b}_j$ are layer parameters shared across all critics. This strategy was also utilized in \citet{niu2020permutation} to learn a family of noise conditioned score functions.
Furthermore a Gaussian kernel-density estimate of the distribution is further used to provide extra regularization to the critics. Recall that for a set of samples $\bfA_{1..N}$ drawn from any distribution, its kernel density estimate is given by:
\begin{displaymath}
p^{KDE}_{h}(\xx) = \sum_{n=1}^N K_h(\xx - \bfA_i)
\end{displaymath}

where $K_h$ is the Gaussian kernel function with bandwidth $h$ i.e $K_h \propto \exp(-(\frac{\xx - \bfA_i}{h})^2)$. For simplicity we set the bandwidth to be the same as $\sigma_i$.

\section{Edgewise Dense Prediction Graph Neural Network}
\citet{niu2020permutation} devised Edgewise Dense Prediction Graph Neural Network (EDPGNN) to learn generative models for graphs. We briefly summarize their architecture here.
EDPGNN extends a standard message passing model with two additions a) EDPGNN have an edge update function along with a node update function and b) EDPGNN is equipped with multiple channels akin to a convolutional network. The message passing and update operation for an EDPGNN with $C$ channels and $T$ message passing steps is given as:

\begin{align*}
  m^{t+1}_{[c,v]} = \bfA^t_{[c,v,w]} h^{t}_w \quad \forall c \in \{0,1,..C\} \\
  h^{t+1}_{v} = \MLP^{\text{Node}}_t( \textbf{CAT}[\tilde{m}^{t+1}_{[c,v]} + (1+\epsilon)m^{t}_w]_{c \in \{0,1,..C\}}) \\
  \bfA^{t+1}_{[c,u,v]} = \textbf{SYM}(\MLP^{\text{Edge}}_t(A^{t}_{[c,u,v]}, h^{t+1}_{u},h^{t+1}_{v})) \\
\end{align*}

where $\textbf{CAT}$ refers to concatenation operator and $\textbf{SYM}$ refers to the symmetrization operator which symmetrizes a matrix. The basic idea is to encode different variations of the same graph in different channels, and compute messages across them. Then for prediction or node updation we aggregate information across channels via flattening the messages into a single vector.

\section{Related Works}
\label{apx:relword_ebm}

\citet{niu2020permutation} proposed a permutation invariant generative model for graphs based on DSM procedure. The architecture of the energy network in our work is the same as the one presented posed by them. The key difference between our work and \citep{niu2020permutation} is two fold. Firstly instead of score matching we have an adversarial training procedure which leads to better samples. Secondly since \citet{niu2020permutation} try to learn the score gradients directly, their model does not provide a density model and cannot be used to assess either the likelihood or unnormalized density of data samples.

\cite{hu2018stein} proposed minimizing the Stein divergence to learn an implicit sampler for a given unnormalized probability model. Their approach backpropagates the gradients of the Stein discrepancy $S$ to the sampler via the reparametrization trick \cite{kingma2013auto}.
\cite{ranganath2016operator} introduced a similar adversarial objective for a different family of critic functions.
Stein divergence based training methods for unnormalized models have also been proposed \citep{grathwohl2020cutting,hu2018stein}. Multiple works \cite{liu2016kernelized, hu2018stein, jitkrittum2019kernel} have used the Stein divergence restricted to RKHS to build goodness of fit tests.

Multiple non-sampling methods train non-normalized models have been proposed in literature. \citet{hyvarinen2007some} proposed a score matching (SM) procedure which minimizes the Fisher divergence. Another method known as Denoising Score Matching (DSM), based on score matching with noise perturbed data was proposed by \citet{vincent2011connection}. Our use of noise conditioned critics has similarities to the use of noise conditioned scores in DSM \citep{song2019generative}. Training of graph EBMs was also recently explored \citep{liu}. However these models relied on MCMC based methods and were not permutation invariant.

Another class of graph generative models is based on variational flows \citep{madhawa2019graphnvp}. Similar to autoregressive models, these are trained via maximum likelihood estimation estimation. While in principle these models can include constraints, in practice these works found that to be detrimental for model performance.

